\pdfoutput=1

\documentclass[11pt]{article}

\usepackage{booktabs}
\usepackage{tcolorbox}
\usepackage[]{ACL2023}
\usepackage{booktabs}
\usepackage{times}
\usepackage{latexsym}
\usepackage{graphicx} 
\usepackage[T1]{fontenc}
\usepackage{amsmath}

\usepackage[utf8]{inputenc}

\usepackage{microtype}

\usepackage{inconsolata}
\usepackage{svg}
\usepackage{hyperref}
\usepackage{tcolorbox}

\title{AI Knowledge Assist: An Automated Approach for the Creation of Knowledge Bases for Conversational AI Agents\\
\texorpdfstring{%
  \raisebox{-0.2\height}{\includegraphics[height=7mm]{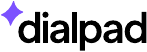}}%
}{}
}

\author{Md Tahmid Rahman Laskar, Julien Bouvier Tremblay \\ \textbf{Xue-Yong Fu}, \textbf{Cheng Chen}, \textbf{Shashi Bhushan TN}\\
          Dialpad Inc. \\
  \texttt{\{tahmid.rahman,julien.bouviertremblay,xue-yong,cchen,sbhushan\}@dialpad.com}\\
 }
\begin{document}
\maketitle
\begin{abstract}
The utilization of conversational AI systems by leveraging Retrieval Augmented Generation (RAG) techniques to solve customer problems
has been on the rise with the rapid progress of Large Language Models (LLMs). However, the absence of a company-specific dedicated knowledge base is a major barrier to the integration of conversational AI systems in contact centers. To this end, we introduce AI Knowledge Assist, a system that extracts knowledge in the form of question–answer (QA) pairs from historical customer‑agent conversations to automatically build a knowledge base.
Fine‑tuning a lightweight LLM on internal data demonstrates state-of-the-art performance, outperforming larger closed-source LLMs. 
More specifically, empirical evaluation on 20 companies demonstrates that the proposed AI Knowledge Assist system that leverages the LLaMA-3.1-8B model can eliminate the cold‑start gap in contact centers by achieving above 90\% accuracy in extracting information‑seeking question-answer pairs from conversations. This enables immediate deployment of RAG-powered chatbots.
\end{abstract}

\definecolor{open_models}{RGB}{237, 242, 242}
\definecolor{closed_models}{RGB}{187, 220, 255}
\section{Introduction}
Generative AI can revolutionize many industries, including the contact center industry\footnote{\url{https://www.salesforce.com/ca/service/contact-center/ai/}}. With the growing demand for high-quality customer service, contact centers are constantly seeking ways to improve their processes \cite{laskar-etal-2023-ai}. One way to achieve this goal is by building conversational agents to help answer customer questions \cite{ferraro2024paradoxes}. Although in real-world scenarios, contact center virtual agents often rely on a comprehensive knowledge base of question-answer (QA) pairs to handle customer inquiries, many enterprises may face a cold start problem if information (e.g., help center articles) related to customer questions are not found in the knowledge base, or if the contact center does not have a knowledge base to begin with \cite{zheng2023dialogqae}. This severely limits the adoption of conversational AI agents in industries. Meanwhile, building a knowledge base from scratch is time-consuming and deters the adoption of such conversational AI systems. 

\begin{figure}
        \centering
        \includegraphics[width=0.7\linewidth,scale = 0.3]{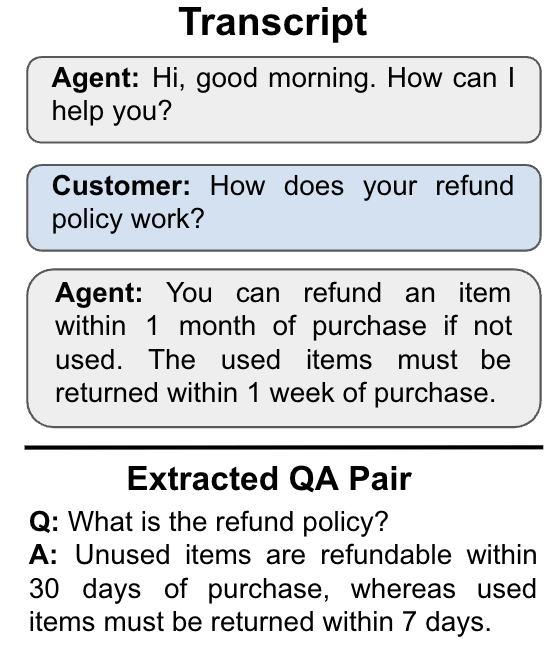}
        \caption{An example of knowledge extracted from transcripts in the form of QA pairs.}
        \label{fig:intro}
\end{figure}

    \begin{figure*}
        \centering
        \includegraphics[width=\linewidth]{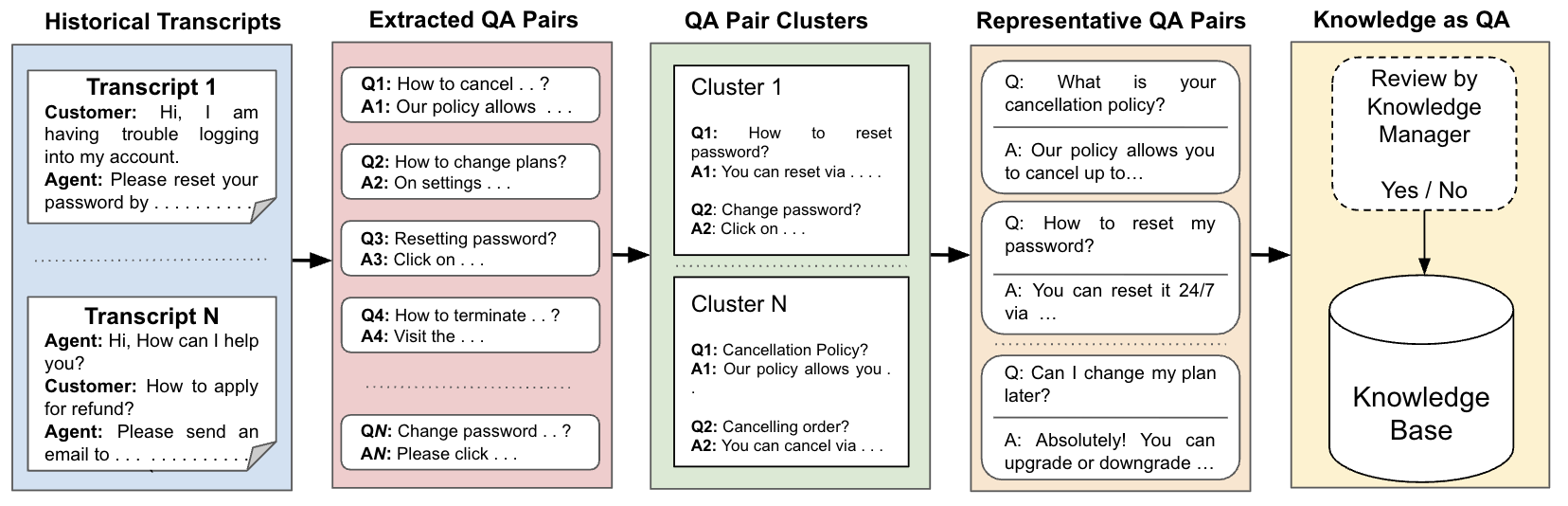}
        \caption{\small{An overview of our proposed AI Knowledge Assist. First, QA pairs are extracted from historical transcripts. Then clustering is applied to group similar QA pairs. Finally, from each cluster, representative QA pairs are constructed and then recommended for the knowledge base (a knowledge manager may review the recommended knowledge before insertion)}}
        \label{fig:overview}
\end{figure*}

Nonetheless, contact centers may possess a wealth of customer service conversation logs (call transcripts and chat histories) that contain repeated information-seeking questions alongside their resolutions \cite{laskar-etal-2023-ai}. Turning past interactions in such historical conversations into an FAQ-style knowledge repository can be useful to develop a knowledge base off the shelf 
\cite{agrawal2024beyond}. This may result in more adoption of the chatbot feature, increasing agent efficiency by handling customer concerns with the help of a dedicated knowledge base, which may ultimately lead to improving customer satisfaction. 

In this paper, we address the cold start problem in conversational AI agents. To this end, we present \texttt{AI Knowledge Assist}, 
a Generative AI-powered system that automatically builds knowledge bases from past conversations. More specifically, we leverage cost-effective LLMs \cite{wan2024efficient} to analyze historical customer-agent conversations to extract knowledge in the form of QA pairs (see Figure \ref{fig:intro} for an example) and save them in a knowledge base to address the cold start problem.
This paper contains a detailed description of our development and evaluation methodology to deploy \textit{AI Knowledge Assist} in real-world contact centers to address customer concerns. Extensive experiments on real-world datasets demonstrate that the proposed \textit{AI Knowledge Assist} system can significantly boost the capabilities of Contact Center AI chatbots to better handle customer concerns.



\section{Related Work}
The recent success of LLMs in zero-shot scenarios in a wide range of tasks \cite{laskar2023systematicchatgpt} has opened the avenue for new application areas in real-world industrial settings \cite{zhang2025evaluating,otani-etal-2025-natural}. This inspires researchers and practitioners to use LLMs in solving complex tasks that require the analysis of noisy conversational transcripts \cite{saini2025llm,zhu2025can,laskar-etal-2023-building,laskar-etal-2024-query}.
Moreover, since LLMs generate human-like responses, the development of conversational AI agents 
is also on the rise\footnote{\url{https://www.genesys.com/definitions/what-is-conversational-ai-for-call-centers}}. 

Nonetheless, prior studies on building conversational AI agents have several limitations: (i) missing discussions on how to tackle the cold start problem when organizations do not have a dedicated knowledge base \cite{agrawal2024beyond,xu2024retrieval}, (ii) requiring human-annotated large training datasets \cite{zheng2023dialogqae} to build models for information extraction from transcripts, which is difficult to obtain in real-world industrial scenarios \cite{fu2022effective}, (iii) limiting the evaluation only on chat logs \cite{zheng2023dialogqae}, ignoring noisy voice transcripts \cite{fu2022effective}.


With prior research demonstrating that LLMs are effective in analyzing noisy conversational transcripts \cite{laskar-etal-2023-building}, in this paper, we propose \textit{AI Knowledge Assist}, a system that leverages LLMs to analyze the call transcripts in contact centers and extracts relevant knowledge from these conversational data in the form of QA pairs. The extracted QA pairs are then stored in a knowledge base to address the cold start problem. Contrary to prior work, our study focuses on addressing the cold start problem in real-world industrial scenarios, with the system being entirely developed in a cost-effective manner from noisy 
transcripts.

\section{Our Proposed Approach}
The \textit{AI Knowledge Assist} system employs a three-stage pipeline, as demonstrated below (also see Figure \ref{fig:overview}).
\subsection{Knowledge Extraction from Transcripts}
The initial step focuses on extracting potential question and answer pairs from historical call transcripts. Given a call transcript, an LLM is prompted to extract information-seeking questions from customers alongside the corresponding answers provided by the agents. Since we utilize voice transcripts, the LLM is also instructed to rewrite the question and the answer instead of mere extraction when needed, such that the QA pairs can be understood without reading the full conversation. The LLM is expected to extract the QA pair as follows:
\begin{equation} \label{eq:qa_extraction}
    \left\{(Q_i, A_i)\right\}_{i=1}^{N(T)} = LLM(T;\theta)
\end{equation}
Here, model parameters are denoted by~$\theta$, which simultaneously extracts and rewrites each QA pair. $N(T)$ is the number of QA pairs that the model finds in the transcript $T$, and $(Q_i, A_i)$ denotes the $i$-th QA pair. In this way, we extract QA pairs from $M$ transcripts (\(T_{1}, T_{2}, \dots, T_{M}\)).
 
\subsection{Clustering for Deduplication}
Once QA pairs are extracted from different transcripts, they may exhibit redundancy (e.g., semantically similar QA pairs may appear in multiple transcripts). If it is not managed, the knowledge base may contain many redundant QA pairs. Therefore, our second step involves clustering these QA pairs into semantically similar groups to facilitate the deduplication and filtering of closely related QA pairs. For this purpose, we first measure the pairwise cosine distance between the question embeddings of every QA pair as follows:
\begin{equation}
    {dist}({q}_i, {q}_j) \;=\; 1 - \frac{{q}_i \cdot {q}_j}{\lVert {q}_i \rVert \,\lVert {q}_j \rVert}
\end{equation}
Here, ${q}_i$ and ${q}_j$ denote the embeddings of the questions in the ${ith}$ and ${jth}$ QA pairs. Finally, a clustering algorithm is applied to group the QA pairs by minimizing the intra-cluster distance and maximizing the inter-cluster distance. 

\subsection{Recommending Representative QA Pairs}
In the final step, we again leverage an LLM to process each cluster of QA pairs. For each cluster, the model selects one or more representative QA pairs that best encapsulate the information in that cluster. 
This step serves a dual purpose: \textit{deduplication} and \textit{filtering}, by ensuring that highly similar questions don't lead to redundant entries; and recommendation, by proposing well-formed informative QA pairs for inclusion in the final knowledge base. The representative QA pairs in the $kth$ cluster can be defined as follows:
\begin{equation}
  \mathcal{R}_k \;=\; LLM\!\bigl(C_k;\,\theta\bigr)
  \label{eq:rep_selection}
\end{equation}
Here, \(C_k\) is the \(k\)-th cluster in the $1,\dots,K$ clusters of QA pairs, \(LLM(\,\cdot\,;\theta)\)
denotes the LLM with parameters~\(\theta\), and \(\mathcal{R}_k\) is the set of
representative QA pairs selected for that cluster. These representative pairs can either be directly inserted into a knowledge base or recommended to a Knowledge Manager for human review before final incorporation into the knowledge base.


 \begin{table*}
\centering
\scriptsize
\begin{tabular}{lcccccccc}
\toprule 
\textbf{Model} & \textbf{Precision} & \textbf{Recall} & \textbf{F1-Score} & \textbf{ROUGE-1} & \textbf{ROUGE-2} & \textbf{ROUGE-L} & \textbf{BERTScore} & \textbf{\# QA Pairs} \\  \midrule
\textbf{Knowledge-Assist-8B-SFT} & \textbf{84.88} & \textbf{84.85} & \textbf{84.86} & 41.26 & 19.68 & 23.87 & 60.12 & 24K \\
\midrule
\textbf{LLaMA-3.1-8B-Instruct} & 58.29 & 57.98 & 58.13 & 42.37 & 18.25 & 25.38 & 60.80 & 24K \\ 
\textbf{DeepSeek-R1-LLaMA-8B} & 51.43 & 48.10 & 49.71 & 39.79 & 15.45 & 22.49 & 58.58 & 21K \\ 
 \midrule
\textbf{GPT-4o-Mini}  & 74.62 & 68.68 & 71.53 & 49.13 & 23.79 & 29.09 & 67.95 & 22K \\ 
\textbf{Gemini-2.0-Flash}  & 82.29 & 60.31 & 69.60 & 47.14 & 24.19 & 28.59 & 62.81 & 18K \\ 
\textbf{Gemini-2.0-Flash-Lite} & 72.30 & 58.81 & 64.86 & 47.07 & 23.70 & 28.81 & 62.09 & 20K \\ 
\textbf{Gemini-2.5-Flash-Lite}   & 76.72 & 70.88 & 73.68 & \textbf{54.17} & \textbf{25.42} & \textbf{28.74} & \textbf{66.86} & 22K \\ 
\bottomrule
\end{tabular}
\caption{\small{Performance in the \textit{Knowledge Extraction from Transcripts} step. Here, `\#' denotes the number of extracted QA pairs.}}
\label{table:results_knowledge_extraction}
\end{table*}

 \begin{table}
\centering
\scriptsize
\setlength{\tabcolsep}{3pt}
\begin{tabular}{lcccc}
\toprule 
\textbf{Model} & \textbf{Precision} & \textbf{Recall} & \textbf{F1-Score} & \textbf{\# QA Pairs} \\  \midrule
{\textbf{Knowledge-Assist-8B-SFT}} & {\textbf{91.4}} & {\textbf{92.2}} & {\textbf{91.8}} & 14K \\
{\textbf{Gemini-2.5-Flash-Lite}} & {81.1} & {78.1} & {79.6} & 13K \\%
\bottomrule 
\end{tabular}
\caption{\small{End-to-End Performance based on 
 \textit{Final Recommendation of Representative QA Pairs}. Here, `\#' denotes the number of  representative QA pairs that are recommended.}}
\label{table:results_knowledge_filtering_recommendation}
\end{table}

\section{Experimental Settings}


\subsection{Dataset} We collected real-world data over a month (November 2024) from contact centers across 20 client companies of Dialpad\footnote{\url{https://www.dialpad.com/}} that consist of customer-agent call conversation transcripts generated using Automatic Speech Recognition systems. On average, each transcript contains 855 words.  
To ensure customer data privacy, the dataset is anonymized using Google Cloud Data Loss Prevention\footnote{\url{https://cloud.google.com/security/products/}} service. 
Note that in real-world settings, obtaining human-annotated data is challenging, which becomes even more difficult in the context of noisy business conversations \cite{laskar2022improving}. Considering these challenges, alongside the customer's data privacy concerns, we annotate our collected dataset using the \emph{Gemini-2.5-Pro}\footnote{\url{https://deepmind.google/models/gemini/pro/}} model by following our proposed approach: (i) QA pair extraction from transcripts using \emph{Gemini-2.5-Pro}, (ii) Clustering the extracted QA pair using the DBSCAN algorithm \cite{schubert2017dbscan} on the question embeddings generated by the BGE-Large\footnote{\url{https://hf.co/BAAI/bge-large-en-v1.5}} \cite{chen-bge-etal-2024-m3} model, and finally (iii) Representative QA pair selection using the \emph{Gemini-2.5-Pro} model.
In this way, we annotate 27500 instances: 12500 for knowledge extraction (5500 for training and 7000 for evaluation) and 15000 for the recommendation of representative QA pairs (2500 for training and 12500 for evaluation). 

\subsection{Model Selection} Since our focus is to deploy this solution in a real-world industrial setting, we select the model to develop the system that can achieve good accuracy with faster inference speed and low cost (see Appendix \ref{cost_analysis} for cost analysis). 
Therefore, by considering the accuracy and efficiency of open-source LLMs in real-world settings \cite{laskar-etal-2023-building,fu-etal-2024-tiny}, we select the LLMs for knowledge extraction and final recommendation that has at least 7B parameters (and does not exceed 10B parameters).  More specifically, we used the LLaMA-3.1-8B \cite{dubey2024llama3} model due to its widespread utilization 
in real-world industrial applications\footnote{\url{https://about.fb.com/news/2025/01/organizations-using-llama-solve-industry-challenges/}}.  For closed-source models, the most cost-effective versions are also preferred. 
 More specifically, we select the mini\footnote{\url{https://openai.com/index/gpt-4o-mini-advancing-cost-efficient-intelligence/}} versions from OpenAI and the Flash\footnote{\url{https://deepmind.google/models/gemini/flash/}} versions from Google's Gemini series models.  For clustering, we use the DBSCAN \cite{schubert2017dbscan} algorithm with BGE-Large \cite{xiao2024c,chen-bge-etal-2024-m3} embeddings since it demonstrates better performance than other approaches (e.g., K-Means \cite{lloyd1982least}) when evaluated in our data (see Appendix \ref{appendix:cluster})

\subsection{Implementation} For the open-source models, we use HuggingFace \cite{wolf2019huggingface} for implementation, and use the respective API providers for the closed-source models.  
For supervised fine-tuning, we use the LLaMA-3.1-8B model. A total of $3$ epochs were run, with the maximum sequence length being set to $8000$ tokens: $4000$ for input and $4000$ for output. The learning rate was tuned between $2e-4$ and $2e-6$ (inclusive). For response generation, we use the default decoding parameters of each model (HuggingFace for open-source, and the official API of OpenAI and Google Gemini for closed-source), but keep the input and output token limits similar to what we use for fine-tuning. All experiments were run on a machine using 8 {NVIDIA A100} GPUs. 

\subsection{Evaluation Settings} 
Since our dataset is annotated by \textit{Gemini-2.5-Pro}, we did not limit the evaluation of our models within reference-wise metrics like ROUGE \cite{lin2004rouge} or BERTScore \cite{zhang2019bertscore} due to the possibility of the presence of bias in our fine-tuned models when compared with Gemini annotations. Inspired by the success of LLMs-as-the-judge \cite{gu2024surveyllmjudge,laskar2024systematic,laskar2025improving}, we also propose the use of an LLM judge for the evaluation of LLM-generated outputs in reference-free settings. To avoid any self-enhancement bias \cite{NEURIPS2023_91f18a12,ye2024justiceprejudicequantifyingbiases} for the models trained using our \textit{Gemini-2.5-Pro} annotated training data, we did not use Gemini series models as the judge. Instead, we use \textit{GPT-4o} \cite{openai2023gpt4} as the judge due to its effectiveness in various evaluation tasks \cite{xiong2025llava}. We specifically instructed the LLM judge to evaluate the following: 

(i) \textit{For the knowledge extraction step}, identify the number of QA pairs extracted correctly from the given transcript by following the rules.

(ii) \textit{For the final recommendation step}, identify the number of representative QA pairs extracted correctly from the given cluster following the rules. 

Based on the above information, we compute the Precision, Recall, and F1 scores. For clustering models' evaluation, we use the Silhouette \cite{rousseeuw1987silhouettes} metric. 

\subsection{Prompt Construction}
To construct prompts for the knowledge extraction step and knowledge recommendation step, as well as for their evaluation using an LLM judge, we conduct extensive prompt engineering on some sampled data to select the best prompt. The selected prompts that we use throughout our experiments can be found in Appendix \ref{prompts}.



\section{Results and Discussions}
In this section, we present our experimental findings. 
We denote our supervised fine-tuned (SFT) model based on LLaMA-3.1-8B as \textbf{Knowledge-Assist-8B-SFT} and compare its performance with various cost-efficient open-source (LLaMA-3.1-8B-Instruct and Deepseek-Distilled-R1-LLaMA-8B) 
and closed-source (GPT-4o-Mini and Gemini-Flash) LLMs. The results of our experiments, as detailed below, highlight the performance of our proposed system in the key stages of knowledge extraction and 
recommendation.
\subsection{Performance on Knowledge Extraction from Transcripts}
As shown in Table \ref{table:results_knowledge_extraction}, our fine-tuned model, {Knowledge-Assist-8B-SFT}, which utilizes the LLaMA-3.1-8B as the backbone, achieves the best performance in the knowledge extraction task in terms of Precision, Recall, and F1-Score, outperforming both closed-source and open-source zero-shot baselines. More specifically, our model achieved an F1-Score of 84.86\%, surpassing GPT-4o-Mini (71.53\%) and Gemini-2.5-Flash-Lite (73.68\%). This indicates the efficacy of fine-tuning on larger LLM-annotated internal datasets for this task. In the reference-wise setting, some closed-source models like Gemini-2.5-Flash-Lite show strong performance in terms of automatic metrics (i.e., ROUGE and BERTScore). However, the references in our evaluation dataset were annotated by the most powerful model in the Gemini series, the Gemini-2.5-Pro model. On the contrary, our Knowledge-Assist-8B-SFT model maintains a competitive edge in the reference-free setting when evaluated by an independent LLM-Judge (i.e., GPT-4o). These findings encourage the use of reference-free metrics in real-world settings to mitigate biases in evaluation datasets when the datasets are annotated using LLMs. 

\begin{figure}
    \centering
    \includegraphics[width=\linewidth]{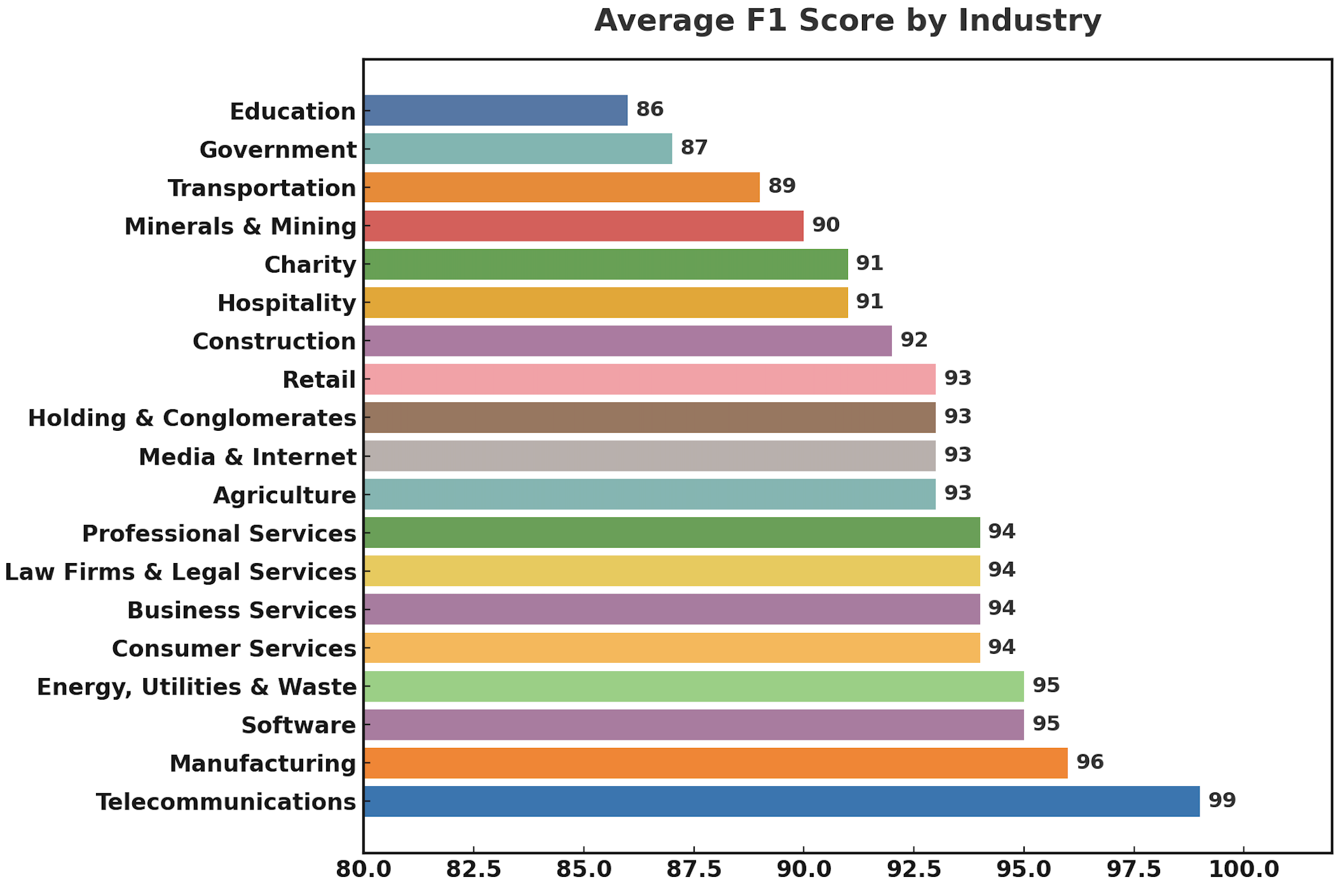}
    \caption{\small{F1-Score per Company type for the Knowledge-Assist-8B-SFT model in terms of the \textit{Final Recommended Representative QA Pairs}.}}
    
    \label{fig:per_company_result}
\end{figure}

\subsection{Clustering Results for Deduplication}
The next step in our \textit{AI Knowledge Assist} system is clustering, where we group similar QA pairs in the same cluster based on the similarity between question-question embeddings. We select the top two models (Knowledge-Assist-8B-SFT model, and Gemini-2.5-Flash-Lite model) from the knowledge extraction step (see Table \ref{table:results_knowledge_extraction}) to group the extracted QA pairs that are semantically similar within the same cluster using DBSCAN \cite{schubert2017dbscan} with BGE-Large Embeddings \cite{chen-bge-etal-2024-m3}. 
From our clustering experiments, we find 1578 clusters for the Gemini-2.5-Flash-Lite model and 1429 clusters for the Knowledge-Assist-8B-SFT model. All the QA pairs in each cluster are then given as input to the LLM to construct the representative QA pairs which are then recommended for insertion in the knowledge base. In the following, we present our findings for the final and most crucial step of our system, recommending representative QA pairs for the knowledge base. 
\subsection{Performance on Recommending Representative QA Pairs}
In this section, we present the end-to-end performance of our \textit{AI Knowledge Assist} system in Table \ref{table:results_knowledge_filtering_recommendation} 
and find that the Knowledge-Assist-8B-SFT model again demonstrates superior performance by achieving an impressive F1-Score of 91.8\%, outperforming the Gemini-2.5-Flash-Lite, which scored 79.6\%. The high precision and recall in the final stage are critical, as they ensure that the knowledge base is populated with accurate and relevant information, directly impacting the performance of the conversational AI agent. Overall accuracy above 90\% 
confirms that our system can effectively bridge the cold start gap for companies. We also find that for the majority of the companies, the F1-Score is also above 90\% (see Figure \ref{fig:per_company_result}). 
\subsection{Impact of Model Choice} 
We conduct some experiments to investigate the importance of model selection:

(i) Does the backbone model choice for fine-tuning impact the performance? 

(ii) Does the annotator model variation impact the performance?

 \begin{table}[t!]
\centering
\scriptsize
\setlength{\tabcolsep}{2.5pt}
\begin{tabular}{lccc}
\toprule 
\textbf{Model} & \textbf{P} & \textbf{R} & \textbf{F1} \\  \midrule
\textbf{Knowledge-Assist-8B-SFT} & \textbf{84.88} & \textbf{84.85} & \textbf{84.86}\\
\textit{ - replace backbone (LLaMA-3.1-8B with Qwen3-8B)}& 77.12  & 72.48 & 74.73 \\
\textit{ - replace annotator (Gemini-2.5-Pro with GPT-4o)} & 80.45  & 55.89 & 65.95 \\ 
\bottomrule
\end{tabular}
\caption{\small{Impact of Model choice on the \textit{Knowledge Extraction} step. Here, `P' and `R' denote `Precision' and `Recall', respectively.}}
\label{table:ablation_studies}
\end{table}

\begin{table}[t!]
\centering
\scriptsize
\begin{tabular}{lcc}
\toprule 
\textbf{Model} & \textbf{Human Preference} & \textbf{Total Approved} \\
 \midrule
\textbf{Knowledge-Assist-8B-SFT} & {25\%} & {107} \\
\textbf{Gemini-2.5-Flash-Lite} & {17\%} & {98} \\
\bottomrule
\end{tabular}
\caption{\small{Results based on Human Evaluation on \textit{Final Recommended QA Pairs}. Here, 58\% of the responses in the preference test were rated as `Tie' by Humans.}}
\label{table:human_evaluation}
\end{table}

Table \ref{table:ablation_studies} presents the results of our experiments. We find that replacing the model from LLaMA-3.1-8B with Qwen3-8B \cite{yang2025qwen3} resulted in a notable drop in all metrics, with the F1-score being dropped to 74.73\%, underscoring the importance of the choice of the base model. Furthermore, when we replaced our data annotator from Gemini-2.5-Pro to GPT-4o, the performance of the model trained on this data decreased significantly to a 65.95\% F1-Score. This underscores the importance of model choice for data annotation. 

\subsection{Human Evaluation} 
We further conduct human evaluations on the final representative QA pairs recommended by \textit{Knowledge-Assist-8B-SFT} and \textit{Gemini-2.5-Flash-Lite}. For human evaluation, we randomly select \textit{100 conversations} to check the following:

(i)  Which model-recommended QA pairs are preferred by humans (\textit{`Tie' if both are preferred})?

    (ii) How many of the final recommended QA pairs are approved by humans (\textit{this mimics real-world scenarios where knowledge base managers would approve the final recommended QA pairs})?

    (iii) What is the agreement between LLM Judge and Human judgments on the final recommended pairs (\textit{we measure the exact match between the number of representative QA pairs recommended for each transcript that are annotated as correct by both the LLM Judge and the Human Annotator})?

This evaluation was conducted by two humans having expertise in data science and computational linguistics. Based on the results presented in Table \ref{table:human_evaluation}, we find that in 25\% cases, Knowledge-Assist-8B-SFT recommended QA pairs are preferred. On the contrary, only in 17\% of cases the Gemini-2.5-Flash-Lite recommended QA pairs are preferred. Moreover, the number of final recommended QA pairs that are accepted by the human evaluators for storing in the knowledge base is 107 for Knowledge-Assist-8B-SFT, while 98 for Gemini-2.5-Flash-Lite. Therefore, the fine-tuned model also demonstrates superiority based on human evaluation, similar to the LLM-judge. 
Furthormore, we find about 90\% agreement between human-annotated judgments and GPT-4o judgments, indicating the reliability of using GPT-4o as the LLM Judge (see Appendix \ref{error_analysis} for some examples of LLM Judge error cases).   We also conducted the Wilcoxon signed-rank test \cite{woolson2007wilcoxon} on the number of approved QA pairs for each model across the 100 samples. With our \textit{knowledge-assist-8B-SFT} model generating a higher number of approved pairs compared to the baseline model, we find that this difference is \texttt{statistically significant} (p $\leq$ 0.05). These findings further provide strong evidence that our system outperforms the baseline in producing content that human evaluators consider correct and valuable for adding to a knowledge base.





\begin{figure}
    \centering
    \includegraphics[width=\linewidth,scale=0.3]{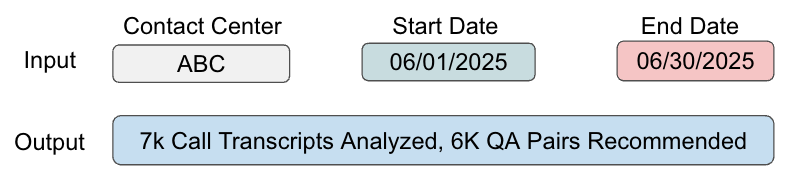}
    \caption{\small{A simple demo of AI Knowledge Assist.}}
     
    \label{fig:simple_demo}
\end{figure}

\section{Real World Deployment and Utilization}
We deploy the \textit{AI Knowledge Assist} system using Kubeflow on the Google Vertex AI Platform\footnote{\url{https://cloud.google.com/vertex-ai}} (requires 1 L4 GPU). 
This setup 
enables automatic execution of the entire system, from data processing to model inference. We show a simple demonstration of \textit{AI Knowledge Assist} in Figure \ref{fig:simple_demo} where the users select the contact center and the timeframe, and then the Kubeflow pipeline automatically analyzes all transcripts in the given timeframe to recommend QA pairs for the knowledge base. 

A key feature of the deployed system should be the ability for the knowledge base to self-update. This can be achieved by continuously processing new call transcripts to extract potential new QA pairs. After applying clustering to construct representative QA pairs 
from the extracted QA pairs, the questions in the representative QA pairs can be compared against the existing questions in the knowledge base by measuring question-question similarity using embeddings. If the similarity score is below a predefined threshold, it indicates a significantly different new customer issue. In such cases, the newly extracted QA pair can be flagged and automatically added to the knowledge base or, for higher-quality control, routed to a Knowledge Manager for review. Moreover, the answer-answer similarity can also be measured for similar questions, and if the similarity score is below a pre-defined threshold, it may indicate that the answer in the knowledge base is obsolete (e.g., changes in the feature/product). This self-updating mechanism ensures that the knowledge base remains current and continuously improves over time by adapting to new customer issues and product changes. 


\section{Conclusion} 
In this paper, we presented
\textit{AI Knowledge Assist}, an LLM-powered system that addresses the cold start problem in conversational AI agents.
 With extensive experiments, we find that our proposed system demonstrates significant effectiveness in automatically creating a knowledge base from historical conversation transcripts. This allows contact centers without an existing knowledge base to still build conversational AI systems by leveraging \textit{AI Knowledge Assist}.  Moreover, we discuss how such systems can be reliably developed, evaluated, and deployed, which can lead to an improved customer experience and agent performance by allowing conversational agents to resolve customer issues more effectively and efficiently by leveraging the constructed knowledge base. 
In the future, we will build new benchmarks to study how to efficiently update existing knowledge bases. 

\section*{Limitations}
As our models are trained on customer-agent conversations, they might not be suitable for use in other domains without further prompt engineering or fine-tuning. Since in this work, we use proprietary data for the development and evaluation of the system, the dataset is not released. However, to maximize methodological reproducibility, we have provided extensive details, including the specific open-source models being used, fine-tuning parameters, and the full verbatim prompts that we used in our experiments.  

\section*{Ethics Statement}

\begin{itemize}

 \item  \textbf{Compensation for Human Evaluation:} Human evaluation was performed by internal scientists who have expertise in computational linguistics. Therefore, no additional compensation was required. Moreover, our in-house employees conducted the human evaluation because of the challenging nature of our proprietary datasets, which contain noisy business conversational transcripts generated by our internal ASR system.\\

 \item  \textbf{Data Privacy:} There is a data retention policy available that allows the user to not grant permission to use their call transcripts for model development. To protect user privacy, sensitive data such as personally identifiable information (e.g., credit card number, phone number) was removed while collecting the data. \\

 \item
 \textbf{License:} We maintained the licensing requirements accordingly while using different tools (e.g., HuggingFace).

\end{itemize}

\bibliography{anthology,custom}
\bibliographystyle{acl_natbib}

\appendix

\section{Cost Analysis}
\label{cost_analysis}

All open-source models in this study have fewer than 10 billion parameters and fit comfortably on a single NVIDIA L4 GPU \cite{laskar-etal-2023-building}.
For proprietary models, Gemini-2.0-Flash-Lite is the most cost-effective at \$0.075 per million tokens for input and \$0.30 for output. Gemini-2.0-Flash costs \$0.15 for input and \$0.60 for output. We omit Gemini-2.5-Flash due to its much higher \$0.30 / \$2.50 pricing. Among the OpenAI models, GPT-4o-mini is modestly higher at \$0.15 / \$0.60, whereas our evaluation model GPT-4o costs \$2.50 / \$10. For data annotation, we rely on Gemini-2.5-Pro at \$1.25 / \$10 per million tokens.

\section{Clustering Model Evaluation}
\label{appendix:cluster}

\begin{figure}[t!]
    \centering
    \includegraphics[scale=0.18]{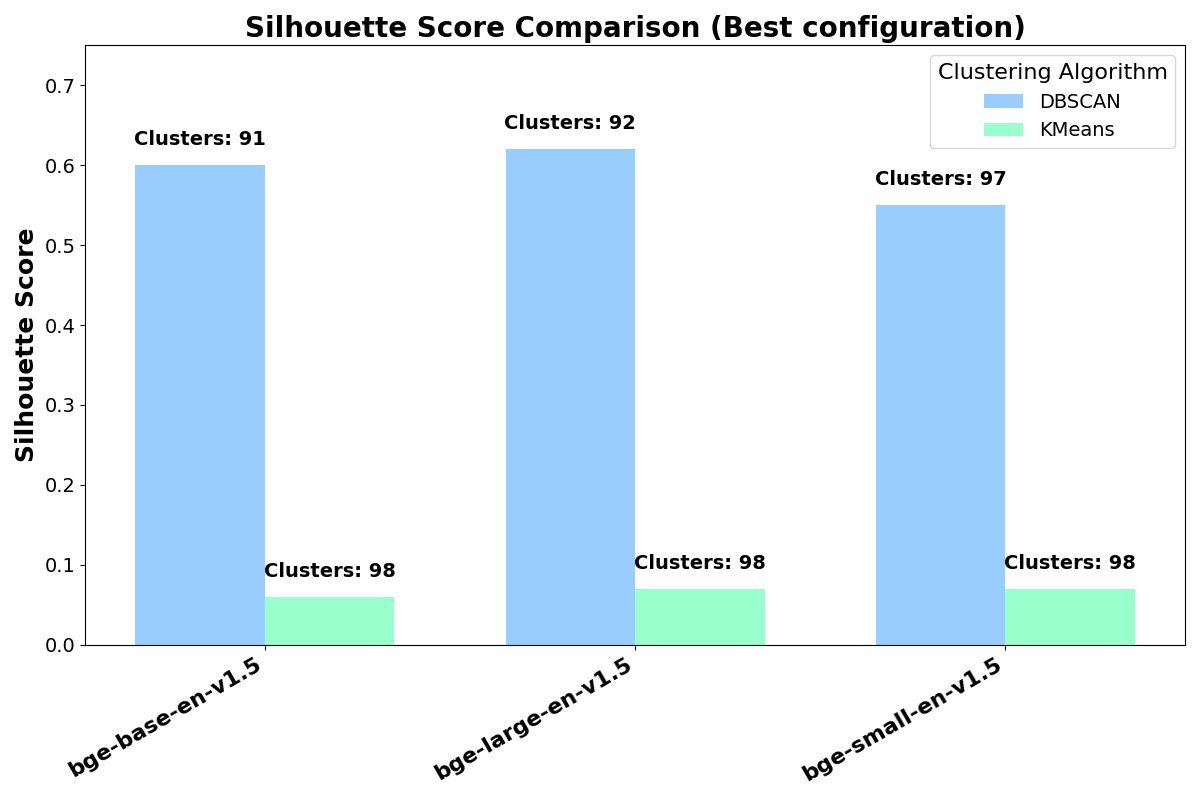}
    \caption{{Performance Comparisons between Clustering Approaches.}}
    \label{fig:cluster_result}
\end{figure}

For clustering, we compare the DBSCAN \cite{schubert2017dbscan} with K-Means \cite{lloyd1982least} using the Silhouette \cite{rousseeuw1987silhouettes} metric. By comparing various embedding models, we find that BGE embeddings \cite{xiao2024c,chen-bge-etal-2024-m3} work best by ensuring both efficiency and accuracy. Based on the results given in Figure \ref{fig:cluster_result} using the best configurations for different BGE embedding models with the minimum sample size of 2 per cluster, we find that the DBSCAN algorithm performs much better than K-Means. Moreover, DBSCAN automatically determines the number of clusters, a significant advantage given the varying call volumes across companies and the high heterogeneity of underlying cluster structures. In contrast, the K-Means method requires a predefined number of clusters \textit{k}, making it unsuitable for datasets whose optimal number of clusters is unknown. 

\section{Sample Prompts}
\label{prompts}

The sample prompts for the LLMs for Knowledge Extraction and Recommendation, as well as for evaluation using LLM Judge, are given below. 

\definecolor{attachedColor}{HTML}{e0efff}
\definecolor{attachedColor2}{HTML}{f3f3f3}
\definecolor{attachedColor3}{HTML}{FFE5CC}
\definecolor{attachedColor4}{HTML}{FFCCCC}
\begin{tcolorbox}[
boxrule=1pt,   
  colback=attachedColor2,    
  colframe=black,           
  colbacktitle=attachedColor, 
  coltitle=black,           
  title={{Prompt: Knowledge Extraction from Transcript using LLMs}},
  fonttitle=\bfseries,      
  fontupper=\small        
]

You are given a call transcript that has two speakers: ``Customer'' (the person seeking help) and ``Agent'' (the customer service representative). \\

Your goal is to extract factually correct high-quality  knowledge from this call in the form of ``questions'' and ``answers'' that will be uploaded to a knowledge base such that if similar questions are asked by a customer in a future conversation, the knowledge base can be used to address the customer concern. \\

Since the transcript may contain stilted or colloquial phrasing due to being transcribed from spoken audio, you can rewrite the knowledge (i.e., QA pair) extracted from the conversation such that the extracted knowledge is human-readable. \\

For knowledge extraction, you should follow these rules:

\begin{enumerate}
    \item Only extract the knowledge that is \emph{non-sensitive} and does \emph{not} contain personally identifiable information (PII).
    \item The extracted knowledge should be \emph{general}, \emph{information-seeking} in nature and applicable to future customers with similar needs. Chit-chats or rapport-building type questions that are not information-seeking should be avoided. 
    \item Do not extract those answers that are \emph{time sensitive}. For instance, if an answer is only applicable till a certain date, ignore such types of knowledge extraction.
    \item If the question is related to any product, then the product name must be mentioned in the selected QA pairs such that they are \emph{understandable} when added to the knowledge base.
\end{enumerate}

\textbf{Output Format}:

 Only return a JSON array of objects without any additional text, where each object has three keys:
 (i) ``Question'', (ii) ``Answer'', and (iii) Justification. \\

In the above, ``Justification'' refers to the rationale behind why the specific ``Question'' and ``Answer'' pairs are selected and how they strictly follow all the rules. In addition, ``Justification'' may also contain some snippets from the conversation transcript that support the extracted knowledge (i.e., the Question-Answer pairs). If no knowledge pair is extracted, just return an empty JSON array.\\

\textbf{Transcript:} 

[Call Conversation Transcript] \\

\end{tcolorbox}

\definecolor{attachedColor}{HTML}{e0efff}
\definecolor{attachedColor2}{HTML}{f3f3f3}
\definecolor{attachedColor3}{HTML}{FFE5CC}
\definecolor{attachedColor4}{HTML}{FFCCCC}
\begin{tcolorbox}[
boxrule=1pt,   
  colback=attachedColor2,    
  colframe=black,           
  colbacktitle=attachedColor, 
  coltitle=black,           
  title={{Prompt: Recommending Representative QA Pairs}},
  fonttitle=\bfseries,      
  fontupper=\scriptsize        
]

You are given a cluster of question-answer (QA) pairs. The cluster is constructed in a way such that similar questions are grouped together in the same cluster. These QA pairs are extracted from different customer-agent conversations and will be stored in a company knowledge base. \\ 


In this task, your goal is to filter the QA pairs in the cluster by constructing representative QA pairs. For this purpose, you should follow the following rules:

\begin{enumerate}
    \item \textbf{No Duplicates}: If there are multiple duplicate QA pairs in the cluster, only extract the QA pair that can be the representative for this cluster. 
    \item \textbf{Rewrite or Extract}: The representative QA pairs can either be extracted directly from the cluster, or you can rewrite them if that makes them better understandable. 
    \item \textbf{Not time-sensitive}: QA pairs that are only applicable for a certain time period (e.g., if something is due today) cannot be representative.
    \item \textbf{Non-personalized}: QA pairs that are specific to certain customers or contain PII information, like account-specific details (e.g., billing info, addresses) cannot be representative.
    \item \textbf{Universal}: QA pairs that are not general and cannot be applicable to future customers with similar needs cannot be representative.
    \item \textbf{Usefulness}: QA pairs that agents cannot use in the company knowledge base to address questions asked by other customers in the future cannot be representative.
    \item \textbf{Information-Seeking}: You should only select the information-seeking QA pairs. Personal questions (e.g., what is your name, address, etc.), chit-chat, or rapport-building type QA pairs should not be included.
    \item \textbf{Understandable}: If the question in the QA pair is related to any product, then the product name must be clearly specified in the representative QA pairs so that it is understandable.
\end{enumerate}

After strictly following the above rules, generate your answer in an array of JSON format, with the following keys:\\
\texttt{(i) Representative Question}\\
\texttt{(ii) Representative Answer}\\
\texttt{(ii) Type}\\
\texttt{(iv) Explanation}\\

Here, the value for \texttt{type} should be either \texttt{"Rewritten''} or \texttt{``Extracted''}, where \texttt{"Rewritten''} means you rewrite it, while \texttt{"Extracted''} means it has been extracted without any rewrite. Moreover, \texttt{Explanation} will contain the reasoning behind \texttt{``rewriting"}, and it should be ``N/A'' if \texttt{``Extracted''}. \\

Note that you should only rewrite in case of urgency. 
For rewriting, you can mix information from multiple questions and answers to create the representative QA pair (if relevant) but ensure that your representative QA pairs do not lose any important information.\\

If similar questions have different answers, you can keep both of them if merging multiple answers into one representative is not possible.\\

The question cluster is given below. Please construct the representative QA pairs.\\

Question Cluster: [List of QA Pairs]

\end{tcolorbox}

\definecolor{attachedColor}{HTML}{e0efff}
\definecolor{attachedColor2}{HTML}{f3f3f3}
\definecolor{attachedColor3}{HTML}{FFE5CC}
\definecolor{attachedColor4}{HTML}{FFCCCC}
\begin{tcolorbox}[
boxrule=1pt,   
  colback=attachedColor2,    
  colframe=black,           
  colbacktitle=attachedColor3, 
  coltitle=black,           
  title={{Prompt: Evaluation of the Knowledge Extraction and Recommendation Models using LLM Judge}},
  fonttitle=\bfseries,      
  fontupper=\scriptsize          
]

You are given the knowledge extracted from a customer-agent conversation in the form of question-answer (QA) pairs that would be stored in the company's knowledge base. This is done so that in a future conversation between a new customer and a new agent in the company, the agent can use the knowledge base to answer the customer's question if similar questions or concerns are asked by the customer.

Your goal is to identify how many of the extracted QA pairs strictly follow the following rules:

\begin{enumerate}
    \item \textbf{Not Time-Sensitive}: QA pairs that are only applicable for a certain time period (e.g., if something is due today) must not be there.
    \item \textbf{Non-personalized and No PII}: QA pairs that are specific to certain customers and contain personally identifiable information (PII) or account-specific details (e.g., billing info, addresses) must not be there.
    \item \textbf{Universal}: QA pairs that are not general and not applicable to future customers with similar needs, such that agents cannot use the QA pairs in the company knowledge base to address questions asked by any new customers, must not be there.
    \item \textbf{Information-Seeking}: Personal questions (e.g., what is your name, address, etc.), chit-chat, or rapport-building type QA pairs must not be there.
    \item \textbf{Customer-Focused}: The selected questions must be something that are asked by the customer, not the agent.
    \item \textbf{Understandable}: If the question is related to any product, then the product name must be mentioned in the selected QA pairs such that they are understandable when added to the knowledge base. 
    \item \textbf{Factually Correct}: In comparison to the conversation context, the QA pair should be factually correct.
\end{enumerate}

Now, your goal is to identify the QA pairs that fulfill the above criteria. For this, you are provided with the Conversation Transcript and the list of QA pairs that are extracted from it. Please generate your answer in the JSON format, with the following keys: \\

(i) \texttt{Total Correct} \\
(ii) \texttt{Total Predicted} \\
(iii) \texttt{Justification} \\

Here, \texttt{Total Correct} denotes the number of correct QA pairs, \texttt{Total Predicted} denotes the total number of QA pairs provided, while \texttt{Justification} denotes your reasoning behind your answer. \\

Transcript: [Call Conversation Transcript] \\

Extracted Questions: [List of QA Pairs]

\end{tcolorbox}

\section{Sample Errors}
\label{error_analysis}

\definecolor{attachedColor}{HTML}{e0efff}
\definecolor{attachedColor2}{HTML}{f3f3f3}
\definecolor{attachedColor3}{HTML}{FFE5CC}
\definecolor{attachedColor4}{HTML}{FFCCCC}
\begin{tcolorbox}[
boxrule=1pt,   
  colback=attachedColor2,    
  colframe=black,           
  colbacktitle=attachedColor4, 
  coltitle=black,           
  title={{Error Examples: LLM Judge Labelling Correct QA Pairs as Incorrect}},
  fonttitle=\bfseries,      
  fontupper=\small          
]
    \texttt{\textbf{Example 1: }} \\\\
    \textbf{Representative Question:} What is [PLATFORM X] and what are its features?\\
    \textbf{Representative Answer:} [PLATFORM X] is a digital platform for administering assessments. It allows users to purchase digital assessment administrations and reports. Users can conduct tests in person, send a remote completion link to clients, or manually enter responses from physical assessments for digital scoring. The platform ensures secure data transmission through encryption and adheres to security guidelines.\\
    \textbf{Type:} Extracted. \\
    \textbf{Explanation:} N/A. \vspace{1mm} \hrule 
\vspace{1mm}
    \texttt{\textbf{Example 2: }} \\\\
    \textbf{Representative Question:} Do customers need to purchase score reports separately when using specific tests or assessments in an online format?\\
    \textbf{Representative Answer:} Yes, score reports need to be purchased separately even if using an online format.\\
    \textbf{Type:} Extracted. \\
    \textbf{Explanation:} N/A. \vspace{1mm} \hrule \vspace{1mm}
    \texttt{\textbf{Example 3: }} \\\\
    \textbf{Representative Question:} What should a customer do if they encounter a problem or technical issue while placing an order online?\\
    \textbf{Representative Answer:} If a customer encounters a problem or technical issue while placing an order online, they have a couple of options. They can try clearing their browser's cache, history, and cookies, and then restarting the browser. Alternatively, they can contact customer service for assistance, and the agent can help them place the order over the phone or manually process it.\\
    \textbf{Type:} Rewritten. \\
    \textbf{Explanation:} The representative question is a combination of several similar questions to encompass all the nuances of the issues. The representative answer synthesizes the recommended solutions from multiple QA pairs, providing both self-service troubleshooting steps (clearing cache) and direct support options (contacting customer service for manual order placement). This approach covers the different solutions offered and is more comprehensive.
\end{tcolorbox}



\end{document}